\pdfoutput=1

\documentclass[11pt]{article}

\usepackage[]{acl}

\usepackage{times}
\usepackage{latexsym}
\usepackage{caption}
\usepackage{graphicx}
\usepackage{subfigure}
\usepackage{amsmath}
\usepackage[T1,T5]{fontenc}
\usepackage[utf8]{inputenc}
\usepackage{microtype}
\usepackage{hyperref}
\usepackage{tablefootnote} 
\usepackage{xcolor}

\title{Composition and Deformance: \\ Measuring Imageability with a Text-to-Image Model}

\author{Si Wu \\ Northeastern University \\  \texttt{siwu@ccs.neu.edu}
\And
        David A. Smith \\ Northeastern University \\ \texttt{dasmith@ccs.neu.edu}}

\begin{document}
\maketitle

\begin{abstract}
Although psycholinguists and psychologists have long studied the tendency of linguistic strings to evoke mental images in hearers or readers, most computational studies have applied this concept of imageability only to isolated words. Using recent developments in text-to-image generation models, such as DALL•E mini, we propose computational methods that use generated images to measure the imageability of both single English words and connected text. We sample text prompts for image generation from three corpora: human-generated image captions, news article sentences, and poem lines. We subject these prompts to different deformances to examine the model’s ability to detect changes in imageability caused by compositional change.  We find high correlation between the proposed computational measures of imageability and human judgments of individual words. We also find the proposed measures more consistently respond to changes in compositionality than baseline approaches. We discuss possible effects of model training and implications for the study of compositionality in text-to-image models.\footnote{Our scripts are available at \url{https://github.com/swsiwu/composition_and_deformance}}

\end{abstract}

\begin{figure*}
\centering
\includegraphics[width=1.0\textwidth]{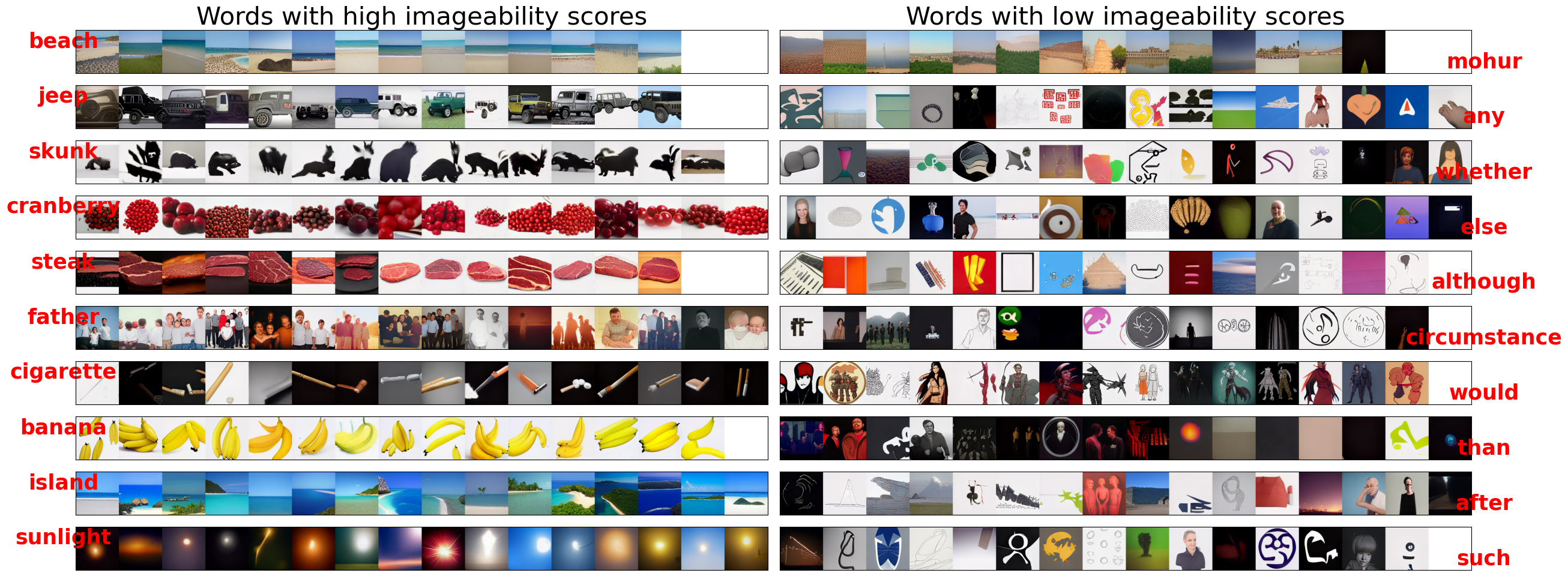}
\caption{Generated images of words with high imageability ratings have more visual homogeneity comparing to the ones with low imageability ratings. }
\label{high-low-MRC-grid}
\end{figure*}

\section{Introduction}
\begin{quote}
Did you ever read one of her Poems backward, because the plunge from the front overturned you? --- Emily Dickinson \cite{samuels}
\end{quote}
\noindent Imageability is the capacity of a linguistic string to elicit imagery. Humans can identify highly imageable words, such as “banana”, “beach”, “sunset”, and words with low imageability, such as “criterion”, “actuality”, “gratitude”; however, it’s difficult to measure imageability computationally. Psycholinguists and psychologists have conducted interviews with humans and released databases of the human imageability ratings, such as the  Medical Research Council (MRC) Psycholinguistic Database, to help researchers in their fields, as well as other fields such as linguistics and computer science, to measure these intangible attributes of verbal content \citep{wilson}. However, conducting these interviews is costly and laborious. Volunteers had to rate hundreds and thousands of words, thus expanding these psycholinguistics databases to the size of modern Natural Language Processing (NLP) corpora such as Corpus of Contemporary American English (COCA), which has more than 60k lemmas with word frequency and part of speech tags, is unrealistic. 

Furthermore, these ratings are only on isolated words. To calculate a sentence’s imageability, many applications have simply added the scores of its component words. Other work uses each word's concreteness level, which research has found highly correlated with imageability \citep{paivio1968concreteness, ellis, Richardson1976ImageabilityAC}. These methods, while they are able to roughly measure imageability, dismiss a fundamental property of a sentence: compositionality. Compositionality depends on word order as well as word choice. Sentences with the same component words but with different word order vary not only in their syntax and semantics, but also their intensity and construction of the imagery, e.g. the famous example: “the dog bit the man” vs. “the man bit the dog”. Previous bag-of-word approaches such as \citet{kao} would consider a sentence and its backward version as having the same imageability, but to human readers, the level of imageability has significantly altered. 

In this paper, we investigate a new computational approach to measure imageability using text-to-image models such as DALL•E mini. We propose two methods to measure the imageability level of both individual words and connected text by generating images with a text-to-image model. We test our methods with both isolated words from the MRC database and connected text from poems, image captions, and news articles, and compare our result with previous bag-of-words methods such as \citet{kao} and \citet{hessel}.  

We propose, firstly, measuring the average CLIP score provided by DALL•E mini and, secondly, calculating the average pairwise cosine similarity between embeddings computed by a pretrained ResNet-18 model. We find that these methods are more highly correlated to human imageability judgments of individual words than other automatic techniques proposed by \citet{hessel}.

We further demonstrate the robustness of our proposed methods by subjecting connected text to various \textbf{deformances}.  As suggested by the epigraph from Emily Dickinson, literary scholars design transformations of the original text to elicit more or less intense reactions from human readers \cite{samuels} and help them calibrate their interpretations of literary works.  This approach is similar to how contrastive training might be used for models such as word2vec or BERT. 

We compare our computational imageability measurements with human judgment collected from Amazon Mechanical Turk (AMT) and found that, on MRC isolated words, our methods have reasonable correlations with human judgment, but on connected text, the correlations vary among different types of connected text. 


\section{Related work}
Paivio introduced the idea of imageability and defined imageability as “the ease/difficulty with which words arose a sensory experience” \citep{paivio1968concreteness, Dellantonio2014ImageabilityNY}. Although imageability is associated with many modalities, some researchers have found that visual modality is its most prominent modality \citep{ellis, Richardson1976ImageabilityAC}. Imageability is also highly correlated with concreteness \citep{paivio1968concreteness, ellis, Richardson1976ImageabilityAC}, and concreteness has also been found to be most related to visual modality \citep{brysbaert}. However, some researchers have found their relation to be more complex: words with high imageability and concreteness “evoke sensations connected to the perception of the objects they denote”, words with high imageability and low concreteness “evoke sensations connected to affective arousal” \citep{Dellantonio2014ImageabilityNY}. An example for the latter is “anger”, which is highly imageable since most of us have the experience of being angry, but “anger” itself is an abstract word. Since imageability and concreteness are highly correlated, in this paper, we will compare some works using concreteness to measure word imageability, but we agree that these two attributes should ideally be disentangled \citep{Richardson1976ImageabilityAC, Boles1983DissociatedIC}.

The imageability rating of an isolated word in psycholinguistics research is usually derived from interviewing human subjects: asking them how imageable a word or a concept is on a 7-point Likert scale \citep{wilson, toglia1978handbook, gilhooly1980age, Coltheart}. Due to the cost of this procedure, the most popular dataset, MRC Psycholinguistics Database, combines three different sources and, even so, has a limited vocabulary size of 9240.

Others have attempted to expand MRC imageability ratings using synonyms and hyponyms identified in WordNet \citep{liu-etal-2014-automatic-expansion}; \citet{Schock2012ImageabilityEF} made a small expansion of 3000 words but only on disyllabic words.

Concreteness ratings in the MRC database were obtained in the same way as imageability ratings, although recently \citet{brysbaert}, using Amazon Mechanical Turk, was able to expand the vocabulary to 37,058 words. 

A concrete idea is assumed to have more shared representation than an abstract idea. \citet{hessel} estimate the concreteness of a word in a database by measuring how clustered a word’s associated images according to the image embeddings provided by ResNet-18. \citet{kastner} estimate imageability using more explicit visual features, such as color distributions, local and global gradient descriptions, and high-level features, such as image theme, content, and composition. However, this approach is supervised and requires a large amount of data to train.

The above works are all on isolated words. Our paper aims to measure a sentence’s imageability beyond bag-of-words methods, where the latter is insensitive to compositional change and imagery loss/gain. We will compare both to the work of \citet{kao}, who measure imageability with bag-of-words models, and of \citet{hessel}, who estimate word concreteness in an unsupervised manner. We will demonstrate our methods’ advantage via correlation with single-word human judgments from MRC (Table~\ref{tab:mrc_img_pearson}). For connected text, we will inspect the measurement change with respect to human expectation (Table~\ref{tab:poem_diff}, Fig~\ref{fig:scatter_percent_change}).

\section{Datasets}
\subsection{Connected text datasets}
\begin{table}[h]
\centering
\begin{tabular}{c | c c} 
 \hline
 Data & imag & concrete\\
 \hline
 Poems & 323.477 & 0.537\\ 
 Captions & 383.270 & 2.659\\
 News & 317.478 & 2.049\\ 
 \hline
\end{tabular}
\caption{Sentence-level average imageability and concreteness scores for different connected text corpora.}
\label{tab:sentence-level-average}
\end{table}
The \textbf{Poetry dataset} consists of 355 English poems written by different types of poets: imagists, contemporary poets, contemporary amateur poets, and 19th-century poets. They were collected by \citet{kao} from different poetry websites and publications: \textit{Des Imagistes} (1914), \textit{Some Imagist Poets} (1915), Contemporary American Poetry (Poulin and Waters, 2006), Amateur Writing (website), and Famous Poets and Poems (website). In their paper, they use various linguistic and psycholinguistic attributes as features for identifying different poets and poem types. In this paper, however, we do not focus on poem-level classification. The dataset was provided by the authors of \citet{kao} for research purposes. 
\\[10pt]
\textbf{Conceptual 12M (CC12M)}\footnote{Available to download at \url{https://github.com/google-research-datasets/conceptual-12m}} \cite{changpinyo2021cc12m} is a dataset of 12 million image-caption pairs, designed for vision-and-language pre-training. We randomly sampled 5000 captions from the dataset. In the 12M captions, real names are replaced with \texttt{<PERSON>}, and some captions contain hashtags. We only use captions with no \texttt{<PERSON>} or \#.
\\[10pt]
\textbf{Cornell Newsroom Dataset}\footnote{Available to download after accepting the data licensing terms \url{https://lil.nlp.cornell.edu/newsroom/download/index.html}} \cite{N18-1065} is a summarization dataset of 1.3 million articles from 38 major English-language news publications. We extract sentences using nltk "sent\_tokenize", then randomly sample 5000 sentences of 10--30 words from the training set original news articles.

\subsection{Psycholinguistics databases}
\textbf{MRC Psycholinguistics Database} contains 150,837 words and their linguistic and psycholinguistic attributes including imageability, concreteness, familiarity, age of acquisition, and Brown word frequency \citep{wilson}. It was originally published by \citet{Coltheart} and made machine-usable by Wilson. The later version also added new entries and made corrections to the previous one. Out of 150,837 words, only 9240 entries have imageability ratings, and there are only 4828 unique words with imageability ratings. The imageability ratings range between 100 and 700. Duplicated imageability word entries are all agreeing on the imageability rating but vary in other attributes, such as different word types (noun, adjective, verb, etc.) and having "N/A" or empty entries. This is possibly because the database was a concatenation of 3 different databases. We will denote this imageability rating as $imageability$ in tables and figures.
\\[10pt]   
\textbf{Brysbaert et al. Concreteness Human Ratings} contains 37,058 English words and 2896 two-word expressions that were crowd-sourced from over 4000 participants on AMT. All lemmas in the dataset were known by at least 85\% of the participants. Concreteness is defined as the ability to have immediate experience through senses or actions and is more experience-based, as opposed to abstractness, which can't be experienced through senses or actions. It's also more language-based. Raters were asked to rate a word on a 5-point scale, where 5 is the most concrete and 1 is more abstract. The Brysbaert ratings are also highly correlated with the MRC Psycholinguistics Database's concreteness ratings, with $r=0.919$. In the following experiments and analysis, we will denote this concreteness rating as $concreteness$.

\section{Methods}
\subsection{Model}
We use DALL•E mini \citep{Dayma_DALL·E_Mini_2021}\footnote{\url{https://github.com/borisdayma/dalle-mini}} as our text-to-image model. DALL•E mini is developed by developers and researchers as an open-source alternative to the original DALL•E developed by OpenAI. It's trained with 15 million webcrawled images and 0.4 billion parameters comparing to the original DALL•E, which is a 12-billion parameter autoregressive transformer trained on 250 million image-text pairs. The image outputs of DALL•E model are ranked by their Contrastive Language-Image Pre-training (CLIP) scores, a neural network that learns to correlate image and text \citep{clip_paper}. Like similarity scores, CLIP score has the range of $[0,1]$; DALL•E mini adjusts this to a percentage in $[0,100]$. 

Specifically, we are using DALL•E mini version "mini-1:v0". One of the hyperparameters for generation is temperature. Temperature acts as a threshold for the quality of the sampled images. We use a temperature of $0.85$ to ensure that the sampled images are highly correlated (high CLIP score) while allowing mild visual diversity. When we did a grid search over this parameter on a small set of poems, it did not have a noticeable effect on the average CLIP scores. Lastly, a higher conditioning scale (cond\textunderscore scale) will result in a better match to prompt but low diversity, and we decided to use a cond\textunderscore scale of 3 (out of 10) informed by a report written by a DALL•E mini developer \citet{Dayma_2022}.

We use 4 Tesla V100 SXM2 GPUs for this paper. For each connected text corpus and each deformance, it takes about 24 hours to generate images. For MRC vocabulary, it takes about 24 hours as well. We will release the code we use for this paper in this GitHub repository\footnote{\url{https://github.com/swsiwu/composition_and_deformance}}.

\subsection{Measurements}

A human can evaluate and “feel” how imageable a text is. For example, “mom is angry at me” is not as imageable as “mom’s eyes are throwing knives at me”. A good computational measurement should be able to quantify and estimate the magnitude of imageability, and when the original text is subject to a compositional change (deformance), it should manifest the direction of change in imageability.

To first examine the text-to-image model’s ability to measure the magnitude of imageability, we will first test on isolated words from MRC and benchmark our methods against the MRC human imageability ratings as well as comparing to other bag-of-words measurements in section~\ref{isolated_word_measurement}. Then in section~\ref{poem_section}, we will test on different connected text. We will alter the original text’s composition and imageability with deformances, and by doing so, we’d like to observe both the magnitude and direction of change using our methods and previous bag-of-words methods. In some deformances, bag-of-words methods will fall short since they don’t consider word order and word choice, while our methods will demonstrate both magnitude and direction of imageability change.

We will also briefly mention how word frequency is unrelated to imageability in section~\ref{word_freq_section}.

\subsection{Measuring isolated word's imageability}\label{isolated_word_measurement}
For the isolated word experiments, our vocabulary is all the words in MRC Psycholinguistics Database that have imageability human ratings. For each word, we will have the MRC imageability rating ($imageability$) and the concreteness rating ($concreteness$) from \citet{brysbaert}. Then we use DALL•E mini to generate a maximum of 16 images for each word to obtain 3 other measurements:
\begin{itemize}
    \item The concreteness score introduced by \citet{hessel}, where each image will only have one label which is the word we used to generate that image, and each word will have a maximum of 16 images associated with that word. We will say Hessel et al. when we refer to this score.
    \item \textbf{Average CLIP score}: our first proposed method. Each image has a CLIP score provided by DALL•E mini when it was generated. We average all generated images' CLIP scores for the target word to produce the average CLIP score. We will denote it as \underline{$aveCLIP$} in tables and figures.
    \item \textbf{Average pairwise image embedding cosine similarity}: our second proposed method. For each generated image, we obtain its image embedding with ResNet-18, then compute the average pairwise cosine similarity score between all images for the target word. We will denote this score as \underline{$imgSim$} in tables and figures. Mathematically, let $M$ be the set of image embeddings, $M= \{\mathbf{m_1}, \mathbf{m_2}, \mathbf{m_3}, ...\}$, $n = |M|$, $N$ be the unique pairs in $M$, and $k = |N|={}_n C_2 $.
    \begin{align*}
        &\text{$imgSim$}= \frac{1}{k} \sum_{(m_x, m_y)\in N} \frac{\mathbf{m_x} \cdot \mathbf{m_y}}{|| \mathbf{m_x}|| ||\mathbf{m_y}||}
    \end{align*}
    This is to be distinguished from Hessel et al.'s method, which calculates the average size of the mutually neighboring images associated with a word and then normalizes it by a random distribution of the image data \citep{hessel}.
\end{itemize}

\begin{figure}[h]
\centering
\includegraphics[width=0.48\textwidth]{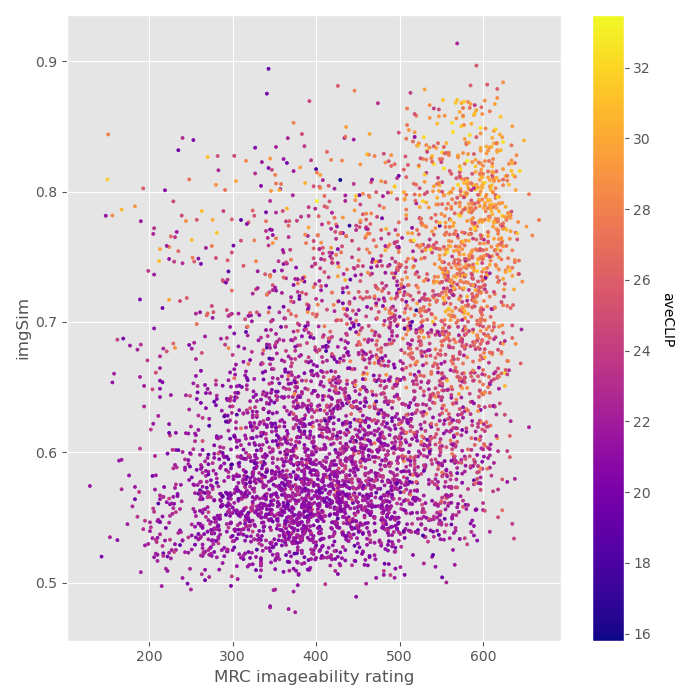}
\caption{X-axis is MRC imageability human rating. Y-axis is $imgSim$, and each dot is a word colored by its $aveCLIP$.}
\label{mrc-figure}
\end{figure}

\begin{figure}[h]
    \centering
    \includegraphics[width=0.4\textwidth]{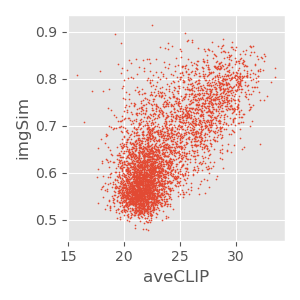}
    \caption{Average CLIP score vs. average pairwise image embedding cosine similarity. Each dot is a MRC word.}
    \label{mrc-cor}
\end{figure}

We visualize these MRC word imageability ratings and their corresponding $aveCLIP$ and $imgSim$  in Figure~\ref{mrc-figure}, where they are colored by  $aveCLIP$. The figure shows that words with very high average CLIP scores tend to have high imageability human ratings and high average image embedding similarity. In Figure~\ref{mrc-cor}, we plot $aveCLIP$ vs. $imgSim$ on the MRC words, and it shows a positive linear correlation between them.

\subsection{The case of familiarity of MRC vocabulary}\label{word_freq_section}
We use word frequency to measure familiarity. Word frequency counts are from Brown Corpus for 3979 out of 4828 MRC words. In table~\ref{tab:mrc_img_pearson}, we show the Pearson Correlation coefficients between all other measurements and MRC imageability ratings. The \citet{brysbaert} concreteness ratings and MRC imageability ratings are highly correlated with $r=0.780$, then followed $aveCLIP$ ($r=0.537$) and  $imgSim$ ($r=0.429$). Word frequency is irrelevant to imageability ratings as it shows a negative and minuscule linear correlation.

\begin{table}[h]
\centering
\begin{tabular}{c c} 
 \hline
 Type & L.C. to imageability ratings \\
 \hline
 word freq & -0.072  \\ 
 concreteness &  0.780  \\
 Hessel et al. & 0.415 \\ 
 aveCLIP & \textbf{0.537} \\
 imgSim & 0.429 \\
 \hline
\end{tabular}
\caption{Linear correlations to MRC imageability ratings.}
\label{tab:mrc_img_pearson}
\end{table}

\section{Connected text and compositionality} \label{poem_section}
\subsection{Preprocessing}
For connected text, the prompt input is each individual caption, news sentence, with the exception that for poems we use every 2 lines (no overlaps) as a single prompt. We use 2 poem lines to ensure enough visual and semantic content for DALL•E mini to generate meaningful images. These two lines are combined with a space character since the majority of the poem lines end with a punctuation mark.

\begin{table*}[h]
\centering
\begin{tabular}{c | c | c} 
 \hline
 \textbf{Deformance} & \textbf{Description} & \textbf{Example}\\
 \hline
  Original &The original poem lines. & “The people pass through the dust On\\
  & &bicycles, in carts, in motor-cars;"\\
  \hline
 Backward &Preserving punctuations and &"Dust the through pass people the\\
         &their locations but reversing&Bicycles on, carts in, motor-cars in;"\\
         &the word order for each line. &\\
\hline
Permuted &Splitting the original sentence by&"The pass people through dust the\\
         &space characters, then randomizing&bicycles, in carts, On motor-cars; in" \\
         &the word order.& \\
\hline
 Just nouns &Keeping only nouns and removing&"people dust bicycles carts motor-cars" \\
            &other words and punctuation. &\\
 \hline
 Replaced nouns &Replacing nouns that can be found&"The \underline{ox} pass through the \underline{murder} On\\
                &in the MRC database with another&bicycles, in carts, in motor-cars;"\\
                &word with the same imageability rat-& \\
                &ings. Plural nouns that can't be found&\\
                &in the database are ignored.& \\
 \hline
\end{tabular}
\caption{Description of different deformances and their examples.}
\label{tab:deformance}
\end{table*}

\subsection{Measuring imageability}
We use the same imageability measurements as the single-word experiments, with these specifications:
\begin{itemize}
    \item Imageability rating: the imageability score for a connected text is the sum of all words' imageability human ratings found in the MRC database divided by the number of words found in the database, as did in \citet{kao}.
    \item Concreteness rating: the sum of all words' concreteness ratings found in the \citet{brysbaert} database divided by the total number of words in the prompt. 
    \item Concreteness score by \citet{hessel}: their method was designed to estimate single word concreteness scores. To get sentence-level concreteness scores, we use the sum of the concreteness scores of all words in a sentence divided by the total number of words. Notice that the same word will have a different concreteness score under a different deformance because a word concreteness score is estimated from all its associated images, and the images are generated using DALL•E mini with deformed sentences as prompt. Also notice that a word concreteness score is not only estimated from one sentence, but all sentences that contain this word under the same type of deformance. We modify their tokenizer so that all punctuation is omitted for a cleaner output. 
    \item Average CLIP score: we average each prompt's generated image CLIP scores, then divide it by the total number of images.
    \item Average pairwise image embedding cosine similarity: we calculate the average pairwise image embedding cosine similarity among the images given a prompt.
\end{itemize}

\definecolor{mypink}{rgb}{0.858, 0.188, 0.478}
\definecolor{myblue}{rgb}{0.0, 0.0, 1.0}
\definecolor{mygreen}{rgb}{0.0, 0.5, 0.0}
\definecolor{myorange}{rgb}{1.0, 0.49, 0.0}

\begin{table*}[h]
\centering
\begin{tabular}{c c | c c c c c} 
 \hline
 \multicolumn{7}{c}{Different imageability measurements' average pairwise percent change compared to the original text} \\
\multicolumn{7}{c}{+: more imageable, -: less imageable}\\
 \hline
 \textbf{Text} & \textbf{Deformance} & \textbf{imag}&\textbf{concreteness}& \textbf{Hessel et al.} & \textbf{aveCLIP} & \textbf{imgSim}\\
 \hline
  Kao \& & \textcolor{mypink}{Backward} & \textcolor{mypink}{0} & \textcolor{mypink}{0} & \textcolor{mypink}{3.842} & \textcolor{mypink}{-0.817} &\textcolor{mypink}{0.046}  \\ 
  Jurafsky & \textcolor{myblue}{Permuted} & \textcolor{myblue}{0} & \textcolor{myblue}{0} & \textcolor{myblue}{490.159} & \textcolor{myblue}{-0.110} &\textcolor{myblue}{-0.182} \\
  Poems & \textcolor{myorange}{Just nouns} & \textcolor{myorange}{24.478} & \textcolor{myorange}{9.644} & \textcolor{myorange}{141.191} & \textcolor{myorange}{1.566} & \textcolor{myorange}{1.253}\\
  & \textcolor{mygreen}{Replaced nouns} & \textcolor{mygreen}{0}  & \textcolor{mygreen}{0.285}  & \textcolor{mygreen}{41.403} & \textcolor{mygreen}{-0.002} &\textcolor{mygreen}{0.130}\\
 \hline
   & \textcolor{mypink}{Backward} & \textcolor{mypink}{0} & \textcolor{mypink}{0} &  \textcolor{mypink}{-4.444} & \textcolor{mypink}{-1.830} &\textcolor{mypink}{-0.886}  \\ 
  Conceptual & \textcolor{myblue}{Permuted} & \textcolor{myblue}{0} & \textcolor{myblue}{0} &  \textcolor{myblue}{111.336} & \textcolor{myblue}{-1.465} &\textcolor{myblue}{-9.398} \\
  12M & \textcolor{myorange}{Just nouns} & \textcolor{myorange}{31.848} & \textcolor{myorange}{16.693} & \textcolor{myorange}{60.494} & \textcolor{myorange}{-0.657} &\textcolor{myorange}{0.049}\\
   & \textcolor{mygreen}{Replaced nouns} & \textcolor{mygreen}{0}  & \textcolor{mygreen}{-1.128}  & \textcolor{mygreen}{8.046} & \textcolor{mygreen}{-3.963} &\textcolor{mygreen}{-3.270}\\
 \hline
  Cornell & \textcolor{mypink}{Backward} & \textcolor{mypink}{0} & \textcolor{mypink}{0} &  \textcolor{mypink}{2.001} & \textcolor{mypink}{-0.757} &\textcolor{mypink}{-0.426}  \\ 
  Newsroom & \textcolor{myblue}{Permuted} & \textcolor{myblue}{0} & \textcolor{myblue}{0} &  \textcolor{myblue}{280.888} & \textcolor{myblue}{-0.899} &\textcolor{myblue}{-0.934} \\
  & \textcolor{myorange}{Just nouns} & \textcolor{myorange}{33.243} & \textcolor{myorange}{3.288} &  \textcolor{myorange}{163.973} & \textcolor{myorange}{1.829} &\textcolor{myorange}{-0.149} \\
  & \textcolor{mygreen}{Replaced nouns} & \textcolor{mygreen}{0} & \textcolor{mygreen}{-0.433}   & \textcolor{mygreen}{176.456} & \textcolor{mygreen}{0.020} &\textcolor{mygreen}{-0.718}\\
 \hline
\end{tabular}
\caption{Comparing different methods percent change between the original and the corresponding deformed text under different deformances.}
\label{tab:poem_diff}
\end{table*}

\begin{figure*}[h]
    \centering
    \includegraphics[width=1.0\textwidth]{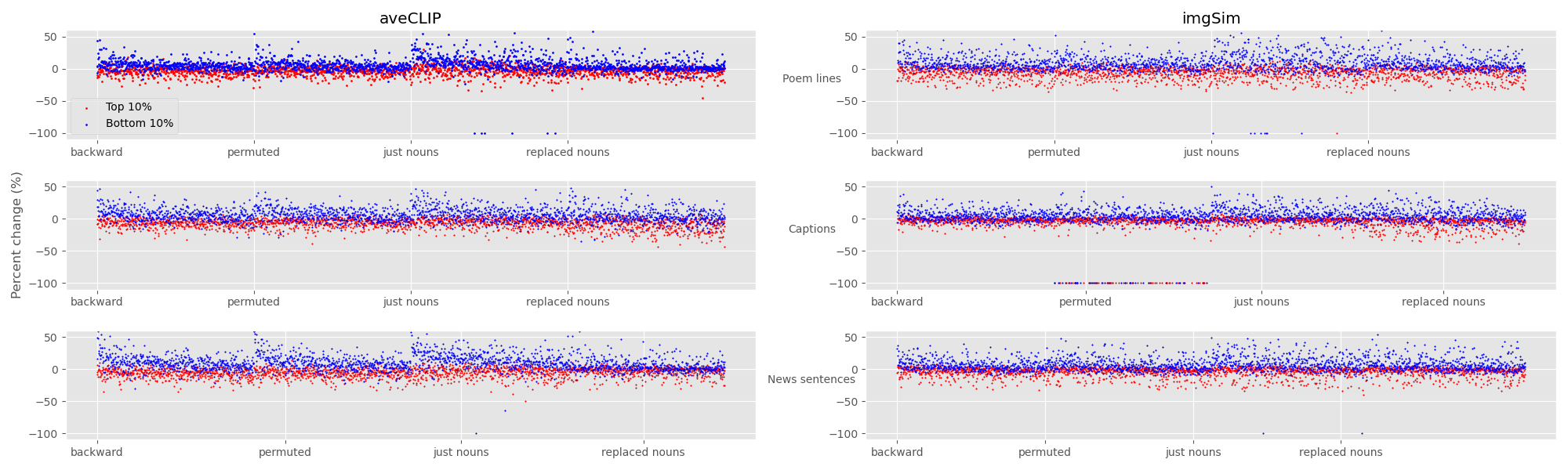}
    \caption{Percent change between lines with the top 10\% and bottom 10\% 
 \{aveCLIP, imgSim\} scores and their associated deformed text.}
    \label{fig:scatter_percent_change}
\end{figure*}

\subsection{Deformances}
The above measurements are repeated for each deformance, and we evaluate the percent change for each measurement. We use percent change instead of difference since these scores are on different scales. A good measurement should show the change in imagery caused by the change of composition. The traditional bag-of-words methods as displayed in Table~\ref{tab:poem_diff} cannot display this change if the component words remain the same; Methods using DALL•E-generated images such as $aveCLIP$, $imgSim$, and Hessel et al. are able to detect changes in both word order and word choice. Hessel et al.'s method, however, does not always correctly show the direction of imagery change.

As defined by \citet{samuels}, a deformance is designed to change text composition by altering its word order and/or word choice. A deformance disturbs the linguistic structure of a sentence, hence it changes not only the surface of the sentence: syntax, word order, and composition of the sentence, but also the underlying information of the sentence: semantics and pragmatics. We perform 4 different types of deformances on each connected text to examine the model’s ability to measure compositional change compared to the bag-of-words methods. The deformances are backward, permuted, just nouns, and replaced nouns, and we provide an example and the elaborated description for each deformance in Table~\ref{tab:deformance}. 

The \underline{backward} and \underline{just nouns} deformances appear in \citet{samuels} \textit{Deformance and Interpretation}, in which they analyze different poetry reading practices. The \underline{backward} deformance alters the word order: even though having the same set of words, it becomes less intelligible. \underline{Permuted} is similar to backward: the dependency structure is disturbed, and it becomes chaotic nonsense. \underline{Just nouns} strips off everything but nouns that are more likely to be imageable, but since there’s no linguistic structure between them, the sentence is less specific in its imagery. Unlike backward and permuted, \underline{replaced nouns} preserves the sentence structure and bag-of-words imageability ratings but alters the imagery via syntax. Backward, permuted, and replaced nouns are experiments where imageability scores remain the same in the bag-of-words approach, but other methods will manifest the change in imagery.  

For replaced nouns, we ignore plural nouns not in the MRC vocabulary. Nouns in both just nouns and replaced nouns deformances are identified by the NLTK tagger.

By construction, all these deformances except \underline{just nouns} cause no change in bag-of-words imageability and concreteness measures. We would expect that applying \underline{backward} and \underline{permuted} deformances to a text would make them less imageable, since the word order becomes less comprehensible, and that is precisely what we see with the $aveCLIP$ and (with one exception) the  $imgSim$ measures. In comparison, the \citet{hessel} metric mostly rates the output of those deformances as far more imageable.

\section{Human judgment}
We recruit workers on Amazon Mechanical Turk (AMT) to rate the imageability of randomly sampled MRC words, poem lines, captions, and news sentences. Workers were informed that they would be participating in a psycholinguistics and natural language processing research study before they accepted the task. For MRC vocabulary, we sample 400 words in total, and for each connected text corpus, we sample 120 sentences for each deformance. We recruit 300 workers in total: the first 100 workers rated 4 MRC words and 6 poem lines each, and the second 100 workers rated 6 captions each, and the rest of the workers rated 6 news sentences. Each worker is only allowed to participate in the entire research once. In total, we recruit 300 workers, and workers are paid \$0.50 for answering 6 or 10 questions. Every participating worker has HIT approval rates for all Requesters' HITs greater than 95\% and number of HITs approved greater than 100, and we require their location to be in the US or Canada. For poems, we mistakenly use two lines of deformed text as one single prompt for workers to rate, and we counter that mistake by averaging the {$aveCLIP$, $imgSim$} from the two lines. Table~\ref{tab:human_judgement_lc} shows the linear correlations between our measurements and the AMT human judgment. We find that while MRC words and captions have relatively high, positive correlations, the linear correlations between poem lines and news sentences are insignificant. We suspect the AMT rating is noisy given that each instance is only judged by one rater. The distribution of human judgments in the appendix also shows interesting variations in rating behavior across corpora.

\begin{table}[h]
\centering
\begin{tabular}{c | c c} 
 \hline
 Type  & aveCLIP & imgSim \\
 \hline
 MRC words  & 0.350 & 0.316 \\ 
 Poem lines & -0.014 & -0.127\\  
 Captions & 0.185 & 0.137 \\
 News sentences  & 0.017 & 0.058\\ 
 \hline
\end{tabular}
\caption{Linear correlations between \{aveCLIP, imgSim\} and human judgment for different corpora across different deformances.}
\label{tab:human_judgement_lc}
\end{table}

\section{Discussion}
Acquiring human imageability judgments is costly and laborious, which makes expanding existing imageability databases difficult. We propose two computational methods that utilize an open-source text-to-image model to estimate isolated words and connected text imageability. Both of our methods require only the input text, and the estimated imageability is calculated based on the properties of the generated images: average CLIP scores and average pairwise image embedding cosine similarity. On isolated words, our proposed methods $aveCLIP$ and $imgSim$ outperform previous unsupervised method proposed by \citet{hessel}: $aveCLIP$ has a linear correlation of 0.537 to MRC human judgment, followed by $imgSim$ 0.429, and Hessel et al. 0.415. Our proposed methods $aveCLIP$ and $imgSim$ also achieve relatively high linear correlations of 0.350 and 0.316 respectively with AMT human judgment, despite the noisiness of collecting that data. 

For connected text, we test our methods on three different corpora: poem lines, captions, and news sentences. Unlike isolated words, sentences' meaning is compositional and depends on word choice and word order. A good sentence imageability method, therefore, should detect the change in imageability caused by compositional change. The biggest downfall of previous bag-of-words methods is that when a sentence is subject to a deformance such as permutation, imageability is unchanged, which is contradictory to human expectation. With a text-to-image model, our methods are able to take the entire sentence as one entity, preserving its composition. We test our methods against a noisy AMT human judgment (Table~\ref{tab:human_judgement_lc}) and obtained vastly different performances on different styles of connected text. We further inspect the performance of these methods by examining the percent change between different deformances and the original text (Table~\ref{tab:poem_diff}). Our methods overall follow human expectation: the imageability level goes down when the original sentence is under a deformance, although we expect our methods to manifest more significant change. In comparison, the direction of change with the method of Hessel et al. doesn't follow our expectation. Although their method can take DALL•E mini generated images as input images, which allows it to learn compositionality from images generated from deformed prompts, it ultimately calculates each sentence's score as a sum of all words in that sentence. Each word’s concreteness score is estimated from multiple sentences of the same deformance that contain that word. We know that a word’s meaning varies in different sentences, thus this method loses a word’s contextual meaning and can not precisely understand compositionality of a sentence.

The language of these three different corpora is very different. Overall, image captions have the highest average imageability rating as well as Brysbaert et al. concreteness rating, with poems being the second most imageable, and news sentences being the second most concrete. Since image captions' language is usually concise, and it possibly has higher noun density, it’s reasonable to see overall the highest impact from deformances under $aveCLIP$ and $imgSim$. All three corpora experience negative impact from permutation under both $aveCLIP$ and $imgSim$, which we assume to be the strongest deformance since it completely randomizes the word order of a sentence. When $aveCLIP$ and $imgSim$ have opposite signs, we notice that a higher absolute value from one measurement also tends to result in a lower absolute value of the other measurement if their signs are different, thus we use Fig~\ref{fig:scatter_percent_change} to explore the percent change distribution between two measurements. In Fig~\ref{fig:scatter_percent_change}, we look at the original lines with the highest and lowest 10\% $aveCLIP$ and $imgSim$ and inspect the percent change between them and their different deformances (Fig~\ref{fig:scatter_percent_change}): the lines with the highest scores consistently decrease their scores after deformances, and vice versa. While the lines with top scores follow our intuition, the increase in lines with low scores reverses the mean: given the lowest score is 0, there isn't much room for the imageability score to fall further.

The performance difference between different connected text also makes us wonder if the training data of the text-to-image model has an effect on the performance. The performance on captions is more contrastive in Table~\ref{tab:poem_diff}, while one of DALL•E mini's training datasets is also Conceptual 12M.

Future work should consider further ways of measuring imageability computationally. As more text-to-image models become available and hopefully more transparent with their training process, we hope researchers will be able to compare different models' performance.

\section*{Limitations}
Since DALL•E mini is trained on English-language material, and since our input text is English only, our proposed methods will only be able to measure the imageability of English isolated words and connected text. 

The text-to-image model we use, DALL•E mini, requires GPUs or TPUs to generate images. While we used 4 GPUs (see section 4 for more details) to obtain the results in this paper, we were able to use a single GPU to successfully run the same experiments with longer runtime.
\subsection{AMT experiments}
We didn't ask the AMT workers what device they were on. Some workers provided feedback via email saying that on mobile phones, the AMT interface didn't show the complete description of the task before they accepted it. Although during the task, detailed instruction was provided, and workers had access to both the brief and long versions of the instruction at any time during the task. It's unclear how the interface will affect the workers' performance and if it would significantly bias their judgment of text imageability.

We were only collecting a single human judgment for each text input. In retrospect, collecting several human ratings per text input and using the average would have reduced noise.

\subsection{Other text-to-image models}
Stable diffusion: using HuggingFace Stable Diffusion release, we generated images using every 2 poem lines as described in section 5.1. The number of generated images per prompt was significantly less than 16, and most prompts generated images that were labeled as harmful even when the prompt didn't have suggestive content. Given this behavior, we decided not to use Stable Diffusion, but we'd like to see future development of Stable Diffusion that allows it to generate abundant and safe images given a prompt.

\section*{Ethics concerns}

Potential risks: DALL•E mini has potential risks of generating offensive images and is vulnerable to other misuses. The poetry corpus we use contains language that might cause DALL•E mini to generate suggestive images. We are concerned about the ethical issues raised by DALL•E mini and similar models and hope further study of DALL•E mini will develop guidelines for responsible use.

\section*{Acknowledgements}
Si Wu was supported by a grant from the Andrew W. Mellon Foundation's Scholarly Communications and Information Technology program. Any views, findings, conclusions, or recommendations expressed do
not necessarily reflect those of the Mellon Foundation. We would like to thank Justine Kao and Dan Jurafsky for providing us with their dataset, and we appreciate all the feedback from anonymous reviewers.

\newpage
\bibliography{anthology, custom}

\appendix
\label{sec:appendix}

\begin{figure*}[h]
\centering
\includegraphics[width=1.0\textwidth]{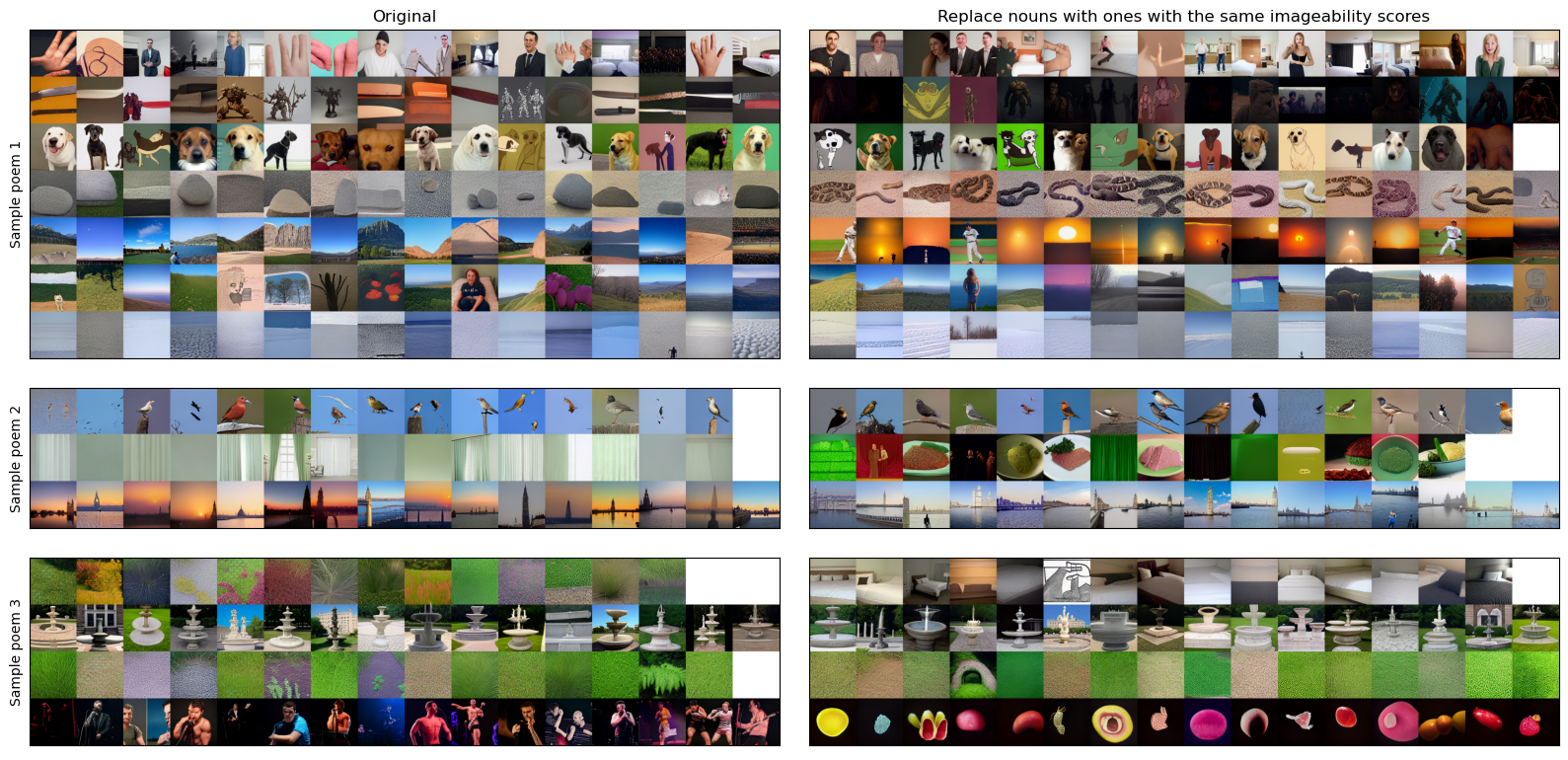}
\caption{The original poem vs. its replaced noun version. Displaying only the changed lines. }
\label{poem-grid-same-noun}
\end{figure*}

\section{A sample of the original poem and its deformed text's generated images (Figure~\ref{poem-grid-same-noun})}

\section{AMT instruction details}
The header: "Please rate the ease or difficulty with which the word/sentence arouses imagery. If an image quickly forms in your mind when reading, give the text a high rating. Only 1 HIT allowed per user."

The short instruction: "Please rate each item from one (low) to seven (high) according to the ease or difficulty with which the item arouses imagery. Any item which, in your estimation, arouses a mental image (i.e., a mental picture, or sound, or other sensory experience) very quickly and easily should be given a high imagery rating; any word/sentence that arouses a mental image with difficulty or not at all should be given a low imagery rating. Please do not go back to refer to your previous ratings."

\begin{figure*}
\centering
\includegraphics[width=1.0\textwidth]{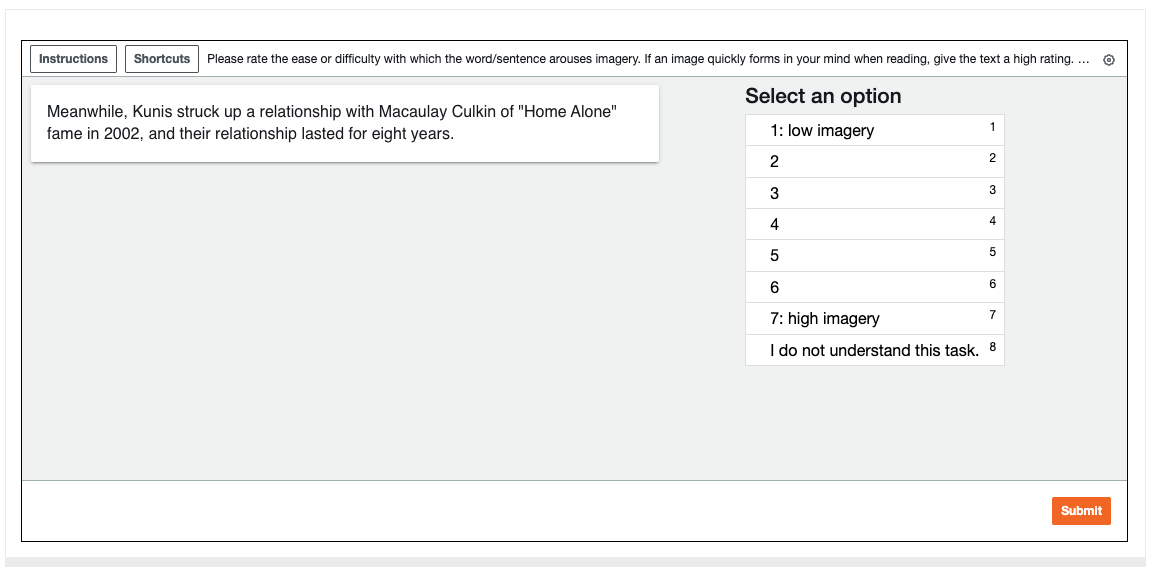}
\caption{A screenshot of the AMT interface that the workers used to participate in our research. The example device was a laptop.}
\label{AMT-interface}
\end{figure*}

\section{A sample screenshot of the AMT interface (Figure~\ref{AMT-interface})}

\begin{figure*}[h]
     \centering
     \begin{minipage}{0.32\textwidth}
         \centering
         \includegraphics[width=\textwidth]{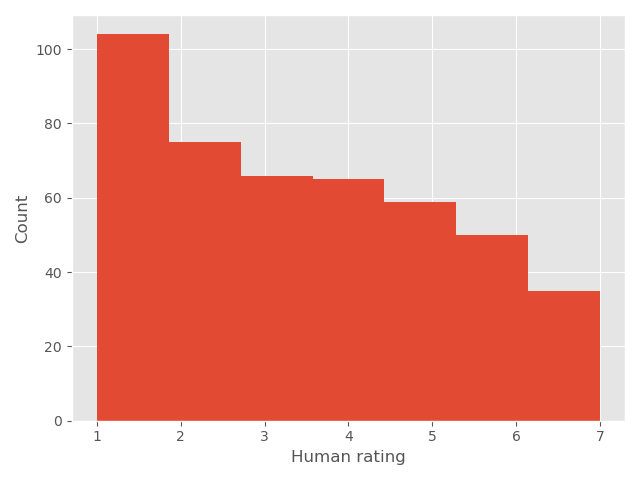}
         \caption{Poem lines}
     \end{minipage}
     \hfill
     \begin{minipage}{0.32\textwidth}
         \centering
         \includegraphics[width=\textwidth]{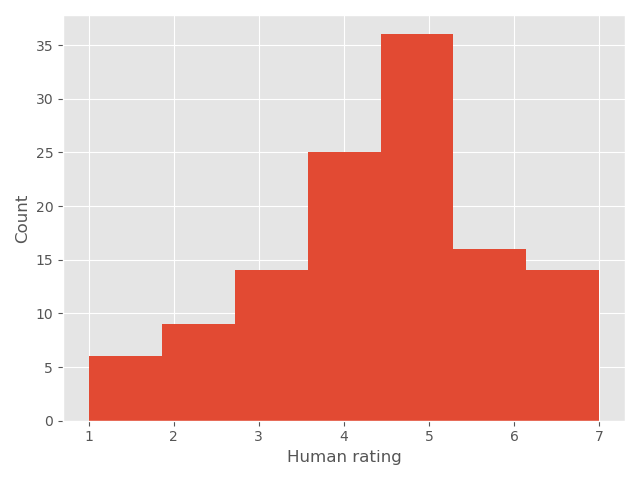}
         \caption{Captions}
     \end{minipage}
     \hfill
     \begin{minipage}{0.32\textwidth}
         \centering
         \includegraphics[width=\textwidth]{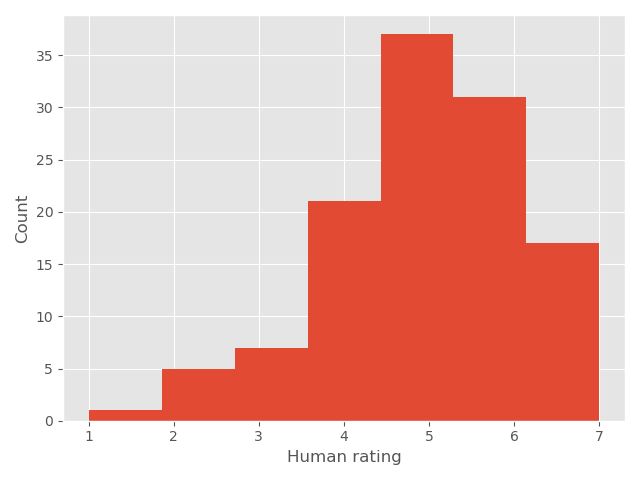}
         \caption{News sentences}
     \end{minipage}
        \caption{AMT human rating distributions for different connected text corpora.}
        \label{fig:human_rating_distribution.}
\end{figure*}

\section{The AMT human rating distributions (Figure~\ref{fig:human_rating_distribution.})}

\end{document}